\documentclass[lettersize,journal]{IEEEtran}
\usepackage{amsmath,amssymb,amsfonts}
\usepackage{algorithmic}
\usepackage{algorithm}
\usepackage{array}
\usepackage[caption=false,font=normalsize,labelfont=sf,textfont=sf]{subfig}
\usepackage{textcomp}
\usepackage{stfloats}
\usepackage{url}
\usepackage{verbatim}
\usepackage{graphicx}
\usepackage{cite}
\usepackage{booktabs}
\usepackage{amsthm}
\usepackage{color}
\usepackage{xcolor}
\usepackage{soul}
\usepackage{multirow}
\usepackage{boldline} 
\usepackage{threeparttable}
\usepackage{hyperref}
\usepackage{colortbl}
\begin{document}
\title{Dsfer-Net: A Deep Supervision and Feature Retrieval Network for Bitemporal Change Detection Using Modern Hopfield Networks}

\author{Shizhen~Chang,~\IEEEmembership{Member,~IEEE,}
        Michael~Kopp,~\IEEEmembership{Member,~IEEE,}
        Pedram~Ghamisi,~\IEEEmembership{Senior Member,~IEEE}
        Bo~Du,~\IEEEmembership{Senior Member,~IEEE}
\IEEEcompsocitemizethanks{\IEEEcompsocthanksitem S. Chang and M. Kopp are with the Institute of Advanced Research in Artificial Intelligence, Landstraßer Hauptstraße 5, 1030 Vienna, Austria (e-mail: szchang@ieee.org; michael.kopp@iarai.ac.at).
\IEEEcompsocthanksitem P. Ghamisi is the Institute of Advanced Research in Artificial Intelligence, Landstraßer Hauptstraße 5, 1030 Vienna, Austria, and also with the Helmholtz-Zentrum Dresden-Rossendorf (HZDR), Helmholtz Institute Freiberg for Resource Technology (HIF), Machine Learning Group, Chemnitzer Str. 40, D-09599 Freiberg, Germany (e-mail: pghamisi@gmail.com).
\IEEEcompsocthanksitem B. Du is with the National Engineering Research Center for Multimedia Software, Institute of Artificial Intelligence, School of Computer Science, Hubei Key Laboratory of Multimedia and Network Communication Engineering, Wuhan University, Wuhan 430079, China (e-mail: dubo@whu.edu.cn).
}}
\markboth{Journal of \LaTeX\ Class Files, 2024}%
{Chang \MakeLowercase{\textit{et al.}}: Dsfer-Net}

\maketitle

\begin{abstract}
Change detection, an essential application for high-resolution remote sensing images, aims to monitor and analyze changes in the land surface over time. Due to the rapid increase in the quantity of high-resolution remote sensing data and the complexity of texture features, several quantitative deep learning-based methods have been proposed. These methods outperform traditional change detection methods by extracting deep features and combining spatial-temporal information. However, reasonable explanations for how deep features improve detection performance are still lacking. In our investigations, we found that modern Hopfield network layers significantly enhance semantic understanding. In this paper, we propose a Deep Supervision and FEature Retrieval network (Dsfer-Net) for bitemporal change detection. Specifically, the highly representative deep features of bitemporal images are jointly extracted through a fully convolutional Siamese network. Based on the sequential geographical information of the bitemporal images, we designed a feature retrieval module to extract difference features and leverage discriminative information in a deeply supervised manner. Additionally, we observed that the deeply supervised feature retrieval module provides explainable evidence of the semantic understanding of the proposed network in its deep layers. Finally, our end-to-end network establishes a novel framework by aggregating retrieved features and feature pairs from different layers. Experiments conducted on three public datasets (LEVIR-CD, WHU-CD, and CDD) confirm the superiority of the proposed Dsfer-Net over other state-of-the-art methods. Compared to the best-performing DSAMNet, Dsfer-Net demonstrates significant improvements, with F1 scores increasing by 4.7\%, 5.9\%, and 2.3\%. Furthermore, when compared to our previous FrNet, Dsfer-Net also achieves noteworthy enhancements, with F1 scores increasing by 2.0\%, 1.4\%, and 4.5\% on three datasets. Code will be available online (https://github.com/ShizhenChang/Dsfer-Net).
\end{abstract}


\section{Introduction}
\IEEEPARstart{R}{emote} sensing (RS) images-based change detection (CD) \cite{jiang2022survey, lv2022land} is intended to identify and analyze landcover variations from multi-temporal images whose geographical locations are correctly matched. This crucial image interpretation task has extensive applications in the analysis of spatiotemporal evolution of the Earth's surface, such as urban planning \cite{prendes2014new}, disaster assessment \cite{zheng2021building}, visual surveillance \cite{carlotto1997detection}, and natural resources management \cite{desclee2006forest}, among others.

In the past few decades, bitemporal change detection has received intensive attention, and myriad methods have been proposed, including traditional methods \cite{touati2019multimodal} and deep learning-based methods \cite{ru2020multi}. Before deep learning became widely used, change detection methods mainly focused on extracting explainable and handcrafted features through the use of statistical analysis \cite{lambin1994change, lv2023spatial}, signal processing \cite{du2019unsupervised, liu2020style}, or machine learning \cite{liu2017change} methods. However, it is difficult to determine which method fits the best in practice, especially for relatively large datasets. Additionally, with the rapid development of high-resolution optical sensors (e.g., WorldView-3, GeoEye-1, QuickBird, and Gaofen-2) \cite{zhang2020deeply}, the growing amount of accessible high-resolution remote sensing images has lead to a number of deep learning-based change detection methods with greater accuracy and more reliable geospatial understanding in complex data analysis.

\begin{figure}
    \centering
    \includegraphics[width=0.95\linewidth]{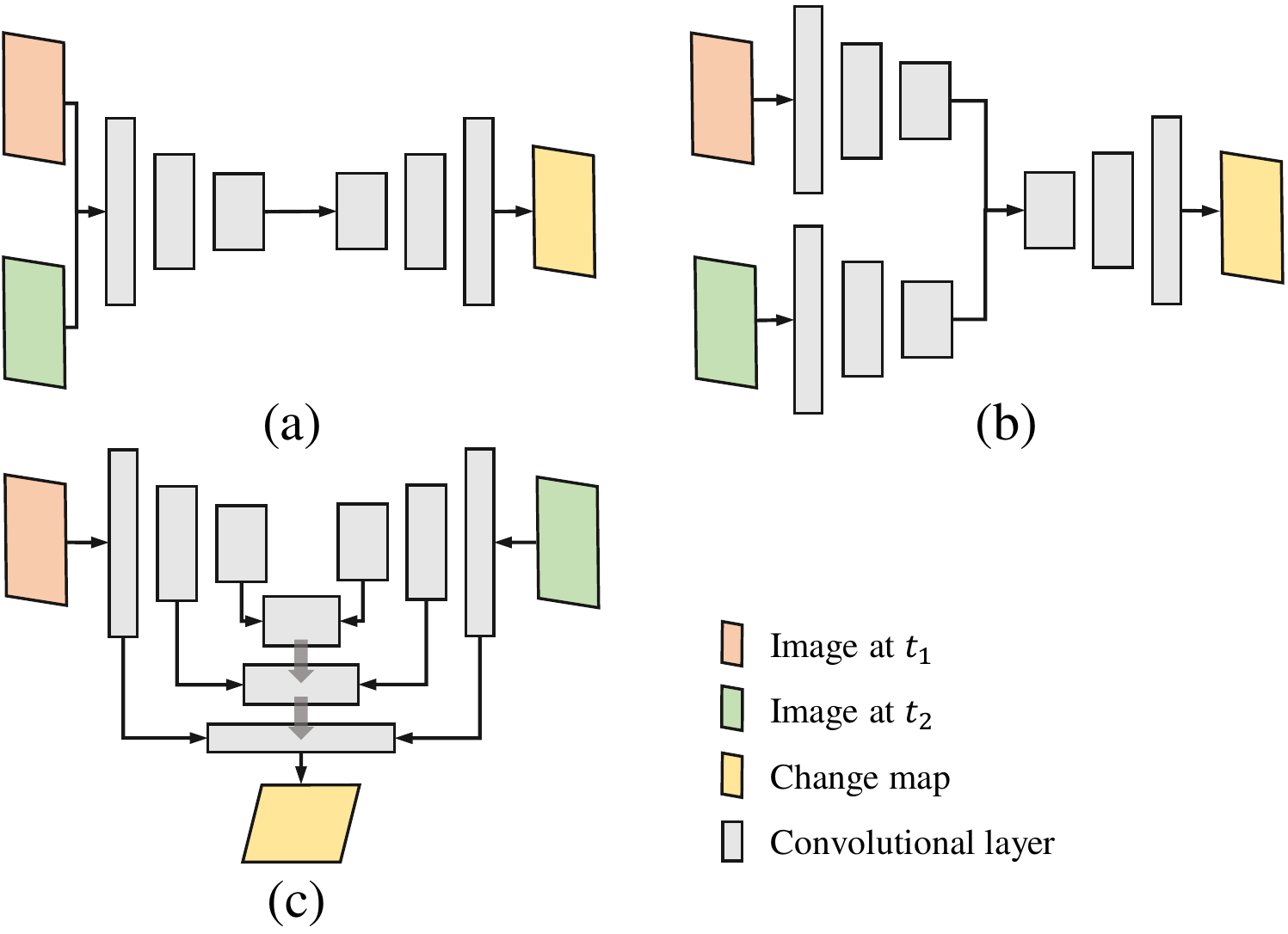}
    \caption{Basic architectures of deep learning-based change detection methods. (a) Single-branch architecture. (b) and (c) Double-branch architectures.}
    \label{fig1}
\end{figure}

 As a powerful architecture of deep learning, convolutional neural networks (CNNs) have been used with increasing frequency due to their strong ability to automatically extract hierarchical multilevel features and learn highly abstract semantic contexts that carry discriminative information about the changing areas. In contrast to single-image processing tasks, designing an appropriate architecture for fusing the image pairs can improve the robustness of feature extraction and change diagnosis \cite{wang2022empirical}. As shown in Fig. \ref{fig1}, there are two main categories of feature fusion: single-branch architecture \cite{lin2022transition} and dual-branch architecture \cite{li2022transunetcd}. The single-branch architecture first fuses the two inputs by image concatenation or calculating the image difference, and feeds them into the network as a whole \cite{peng2020optical}. It utilizes the early fusion (EF) strategy and then converts the fused image into high-level features. However, this approach only fuses partial information and cannot fully consider individual information \cite{lei2019multiscale}. The dual-branch architecture is a late fusion strategy that first learns the single-temporal features, and then merges the learned features as the input of the decoder. Essentially, a Siamese neural network with shared weights has fewer parameters and faster convergence. Although the dual-branch architecture outperforms the single-branch image-level fusion approach, it is inevitably limited by the fact that the shallower layers cannot effectively learn useful features due to the vanishing gradient phenomena. Therefore, making a reasonable combination of the features from both deeper and less deep layers is more helpful to effectively identify the features of the original images that give rise to changes.

\begin{figure}
    \centering
    \includegraphics[width=0.99\linewidth]{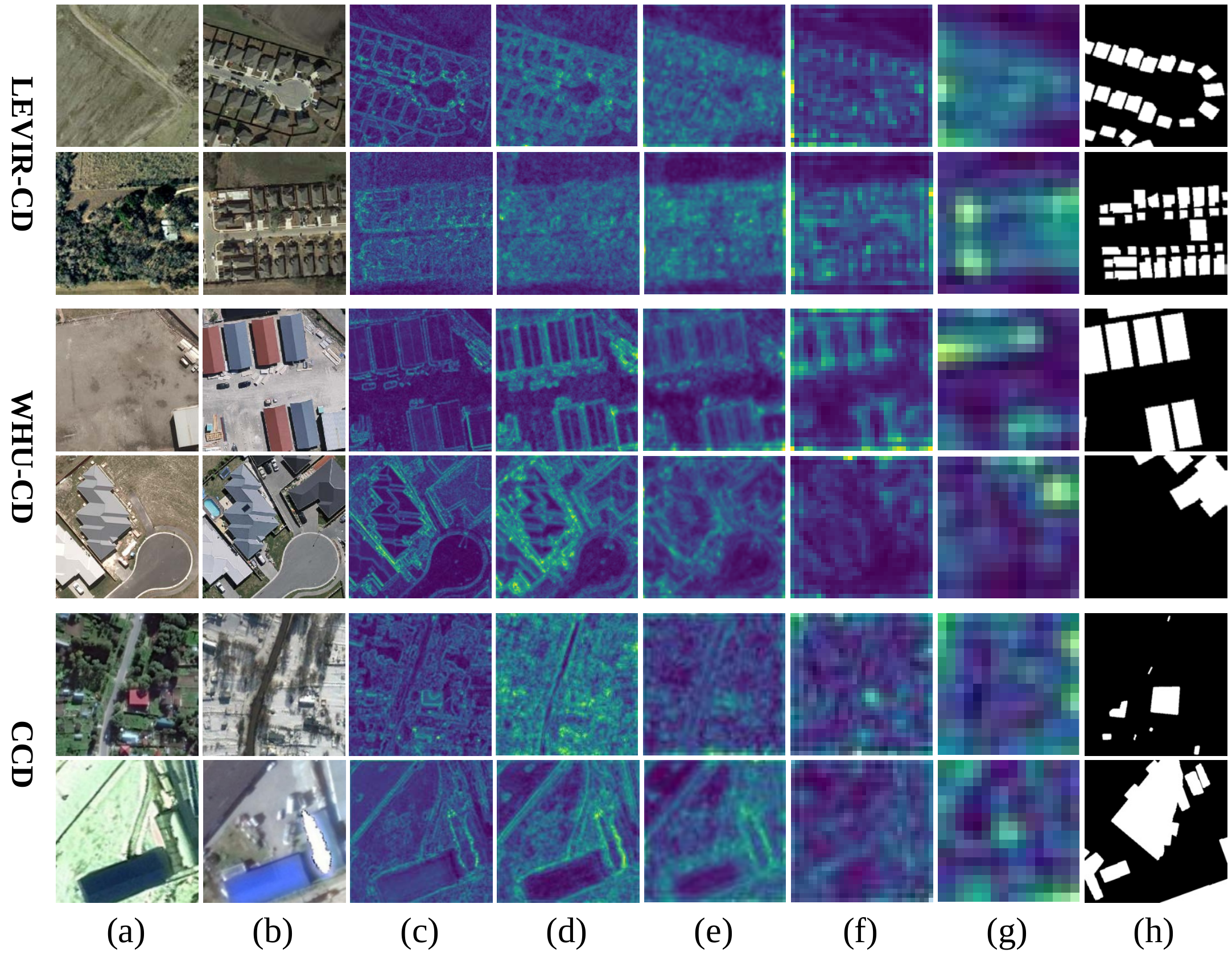}
    \caption{Examples of the difference features from shallow to deep layers in a Siamese change detection network using VGG-16 as the backbone. (a) Images from $t_1$. (b) Images from $t_2$. (c) Features from the first stage. (d) Features from the second stage. (e) Features from the third stage. (f) Features from the fourth stage. (g) Features from the fifth stage. (h) Ground-truths.}
    \label{fig2}
\end{figure}

To solve the aforementioned issues and accurately extract features, deeper and more complex CNN-based networks have been designed. It is well understood that features at shallow layers primarily encapsulate the spatial structure of the image, including texture, boundary, and colors, whereas those at deep layers encompass more abstract content and representative semantic concepts \cite{yosinski2015understanding}. Consequently, multi-scale fusion strategies (see Fig. \ref{fig1}(c) as a visual representation) have been utilized to integrate features from different layers. Several enhanced networks with enlarged receptive fields have been proposed, often incorporating other architecture components such as generative adversarial network (GAN) \cite{hou2019w}, Long Short-Term Memory (LSTM) \cite{sun2020unet}, and spatial and channel attention mechanisms \cite{zhou2022spatial}. However, simply fusing features learned by Siamese architecture with very deep layers may not necessarily improve the performance of change detection methods, as it often results in an explosion of parameters and computational memory requirements. In Fig. \ref{fig2}, we observe the performance of a Siamese network utilizing VGG-16 as the backbone. Notably, the preservation of features relevant to the changed area is optimal in mid-level layers. While shallow layers retain rich texture details, they struggle to filter out information from the unchanged area. Conversely, high-level features, due to repeated pooling, tend to become overly abstract, hindering the network's ability to accurately identify changes of interest. Moreover, the architectural constraints of Convolutional Neural Networks (CNNs) dictate that features from higher levels are intrinsically linked to the shallower layers of the corresponding branch, thus limiting the exploration of temporal information between image pairs.

To address these challenges, we propose a deep supervision and feature retrieval network, abbreviated as Dsfer-Net, for bitemporal change detection. To enhance the efficacy of deep feature extraction in the context of bitemporal change detection, offering improved performance and accuracy in identifying spatial and semantic changes across time, our framework incorporates a Deeply Supervised Feature Retrieval (DSFR) module inspired by modern Hopfield networks (MHNs) in deep learning. This module leverages dual Hopfield layers to aggregate space-time features, facilitating a sequential understanding and tracking of changes within image pairs across two streams. Specifically, deep supervision is integrated into the DSFR module to ensure the comprehensive restoration and retrieval of both semantic concepts and spatial structures within high-level features. The retrieval operation aims to enhance the efficacy of deep feature extraction in the context of bitemporal change detection, offering improved performance and accuracy in identifying spatial and semantic changes across time. A decoder is then employed to decode concatenated feature pairs, employing a multi-scale fusion strategy to achieve precise localization by connecting them with the retrieved features. Finally, to optimize training, we employ a hybrid loss function aimed at minimizing errors in network propagation and feature retrieval.

The main contributions of this article are summarized as follows:
\begin{enumerate}
\item We propose a novel change detection network that combines deep supervision and a retrieval concept for bitemporal remote sensing images. Here, we make further attempts to introduce Hopfield layers for change detection and explain the efficacy of deep feature retrieval and aggregation. The proposed network employs multi-scale fusion with a weighting strategy through a Siamese architecture to comprehensively capture the change of interests from shallow to deep layers.
\item To help high-level features preserve the changed information while generalizing semantic content, DSFR modules are designed for deep layers. It interprets the global context from a sequential understanding and can retrieve differences while the features are highly abstract. A hybrid loss is added to balance the error of the entire network and the Hopfield block.
\item Extensive experiments demonstrates that the proposed Dsfer-Net has superior performance in maintaining the shape and preserving the boundaries of changed objects. In a comparison of different datasets, the proposed model shows significant superiority over other state-of-the-art models.
\end{enumerate}

The rest of the article is structured as follows. Section II reviews related work. Section III introduces the proposed Dsfer-Net method. Section IV presents experiments and discusses the findings. Finally, the conclusions are outlined in Section V.

\section{Related Work}
\subsection{Deep Learning-Based CD Methods}
According to various analyses of the purposes of change detection, deep learning-based CD methods can be categorized into three classes: pixel-based methods, object-based methods, and deep feature-based methods \cite{sun2021iterative}. Pixel-based methods usually apply a deep neural network to transform the images into deep feature spaces, and then compare the features pixel by pixel to obtain the desired change output. For example, Zhang et al. \cite{zhang2016feature} utilize a deep belief network (DBN) to capture key information for discrimination. Then, the bi-temporal features are mapped into a 2-D polar domain to characterize the change information, and the changing map is obtained by adopting a supervised clustering algorithm. Du et al. \cite{du2019unsupervised} design a deep slow feature analysis module that first projects the input images into the deep feature space, then calculates the feature differences using the slow feature analysis metric, and obtains the change map. Chen et al. \cite{chen2020dasnet} utilize pre-trained VGG-16 architectures and attention mechanisms to extract features; the change map is ultimately derived by calculating the weighted double-margin contrast loss.

The main idea of object-based methods is to extract features by segmenting input images and then identifying changes in the state of the segmented objects \cite{lei2019multiscale}. Ji et al. \cite{ji2019building} present a hybrid CNN-based change detection framework where buildings are viewed as objects and extracted through a mask R-CNN. Zhang et al. \cite{zhang2021escnet} proposed an end-to-end superpixel-enhanced CD network that adopts superpixel sampling networks to learn task-specific superpixels and uses the UNet-based Siamese network to exploit multi-level change clues. Liu et al. \cite{liu2021change} propose an object-based change detection framework that utilizes CNN to extract features of image patches and combines different fusion strategies to discriminate changes. We conclude that for both pixel-based and object-based change detection methods, deep learning modules such as DBN, LSTM, attention mechanisms, and CNNs are utilized as tools to transform raw images into high-level features, retaining the most discriminative information related to the change of interests and suppressing the unchanged regions. 

Due to their excellent performance in feature extraction, CNN-based deep learning networks have been widely used in high-resolution remote sensing change detection tasks. To extract more reliable information from the available labeled data, end-to-end deep learning architectures, the fully convolutional neural networks (FCN), have been developed for deep feature-based change detection. In \cite{daudt2018fully}, Daudt et al. present three FCNs based on the UNet model: FC-EF, FC-Siam-conc, and FC-Siam-diff. The FC-EF method adopts an early fusion strategy that concatenates the image pairs as one input of the network; FC-Siam-conc and FC-Siam-diff are Siamese extensions of the FC-EF. The encoding layers of FC-Siam-conc and FC-Siam-diff are separated into two streams of equal structure with shared weights, as in a traditional Siamese network. The difference between these two models is that FC-Siam-conc concatenates the two skip connections in the decoders, while FC-Siam-diff calculates the absolute value of feature differences and concatenates them with skip connections in the decoding stage. A similar architecture was simultaneously proposed in \cite{chen2018mfcnet}. In a later study \cite{daudt2019multitask}, Daudt et al. modified the FC-EF architecture to use residual blocks with skip connections to improve the spatial accuracy of the result. Peng et al. \cite{peng2019end} proposed an improved UNet++ network that first concatenates co-registered image pairs as input, and then adopts the fusion strategy of multiple side outputs to combine change maps from different semantic levels.

While these methods demonstrate good change detection capabilities, they fail to fully incorporate spatial context information and internal relationships from shallow to deep layers, leaving them susceptible to pseudo-changes induced by factors such as noise, angle, shadow, and context. Consequently, researchers have proposed enhanced algorithms aimed at addressing these shortcomings. Furthermore, these methods leverage temporal relationships between input data to produce more robust outputs. Sun et al. \cite{sun2020unet} integrate LSTM into the UNet architecture to use the convolutional LSTM as the core convolutional structure. This method combines convolution and recurrent structure and proves that RNNs are effective for capturing temporal relationships between images. Tang et al. \cite{tang2023object} present a frequency decoupling interaction (FDINet) method for object fine-grained change detection (OFCD), introducing a wavelet interaction module (WIM) for spatial-temporal correlation. Wang et al. \cite{wang2023sscfnet} employed a spatial-spectral cross-fusion module that executes various convolution operations to enhance semantic features. Lei et al. \cite{lei2023ultralightweight} introduced an efficient ultralightweight spatial-spectral feature cooperation network (USSFC-Net), incorporating a spatial-spectral feature cooperation strategy that directly learns 3-D spatial-spectral attention weights without introducing additional parameters.

Recently, attention mechanisms have shown great potential in the field of image processing: they can automatically enhance features based on correlation among pixels within the feature itself. Peng et al. \cite{peng2020optical} present a Difference-enhancement Dense-attention Convolutional Neural Network (DDCNN) that consists of several up-sampling attention units to model the internal correlation between high-level and low-level features. Patil et al. \cite{patil2021effcdnet} develop an optimized architecture that adopts a Siamese-based pre-trained encoder with an attention-based UNet decoder to reconstruct fine-grained feature maps. Lv et al. \cite{lv2022spatial} combine spatial-spectral attention mechanism and multi-scale dilation convolution modules to capture changes with a greater degree of positivity. Zhao et al. \cite{zhao2023high} propose a change detection method based on feature interaction and multitask learning (FMCD) which adopts the Mix Attention Block (MAB) to improve the model's sensitivity to changes. In contrast to conventional CNNs used as feature extractors, recent advancements in change detection algorithms have introduced transformer-based approaches. For instance, GeoFormer, proposed by \cite{zhao2023geoformer}, incorporates a nonlocal Siamese encoder that merges geometric convolution with a transformer. This innovative combination enhances the model's ability to provide local geometric representation information for remote contextual features.

The structure of the attention mechanism places a limitation in that the features can only be enhanced on a specific scale (channel, spatial, or positional), while ignoring the sequential relationships between the two streams. Thus, inspired by MHN, we introduce the Deeply Supervised Feature Retrieval (DSFR) module into the change detection framework.

\subsection{Image Retrieval and Feature Aggregation}
The content-based image retrieval aims to search for relevant images, given query image, from a large-scale database by analyzing the color, texture, or shape of the contents \cite{chen2021deep}. The exponential growth of visual data has fueled rapid development in content-based image retrieval, where relevant images are retrieved from a large-scale database based on the analysis of color, texture, or shape \cite{chen2021deep}. Content-based image retrieval, encompassing instance and semantic tasks \cite{barz2021content}, necessitates accurate and efficient retrieval of discriminative feature representations across large-scale image collections. 

Over the past two decades, image retrieval has undergone astonishing progress with the emergence of deep learning for feature extraction. Prior to the breakthrough of deep learning, handcrafted features such as Bag of visual Words (BoW) \cite{sivic2003video} and Scale-Invariant Feature Transform (SIFT) \cite{lowe2004distinctive} were commonly used. Deep learning,capable of learning multi-level feature representations directly from data, has since demonstrated indisputably superior performance in this field. Convolutional layers in deep learning models are known to capture structure details connected only to local regions of the image with a certain receptive field. To enhance the discriminability of activations extracted from deep CNNs and help retrieve specific instances, feature embedding and aggregation strategies have been introduced to image retrieval. In \cite{noh2017large}, Noh et al. present an attentive local feature descriptor to select keypoint features based on convolutional neural networks for large-scale landmark retrieval. Chaudhuri et al. propose a Siamese graph convolutional network that uses contrastive loss to minimize the most discriminative embedding space for remote sensing image retrieval \cite{chaudhuri2019siamese}. Other aggregation metrics have been combined with off-the-shelf deep learning models for image retrieval, such as Vector of Locally Aggregated Descriptors (VLAD) \cite{gong2014multi}, Selective Match Kernel (SMK) \cite{yang2018dynamic}, deep hashing embedding \cite{do2017simultaneous}, among others.

Unlike previously mentioned image retrieval systems, our work introduces a DSFR module capable of directly executing the retrieval process within the deep convolutional layers. It's important to note that this module retrieves deep differential features under supervision using a modern Hopfield network and aggregates the retrieved feature pairs. These pairs are then forwarded for bitemporal remote sensing change detection.
\begin{figure*}
    \centering
    \includegraphics[width=0.93\linewidth]{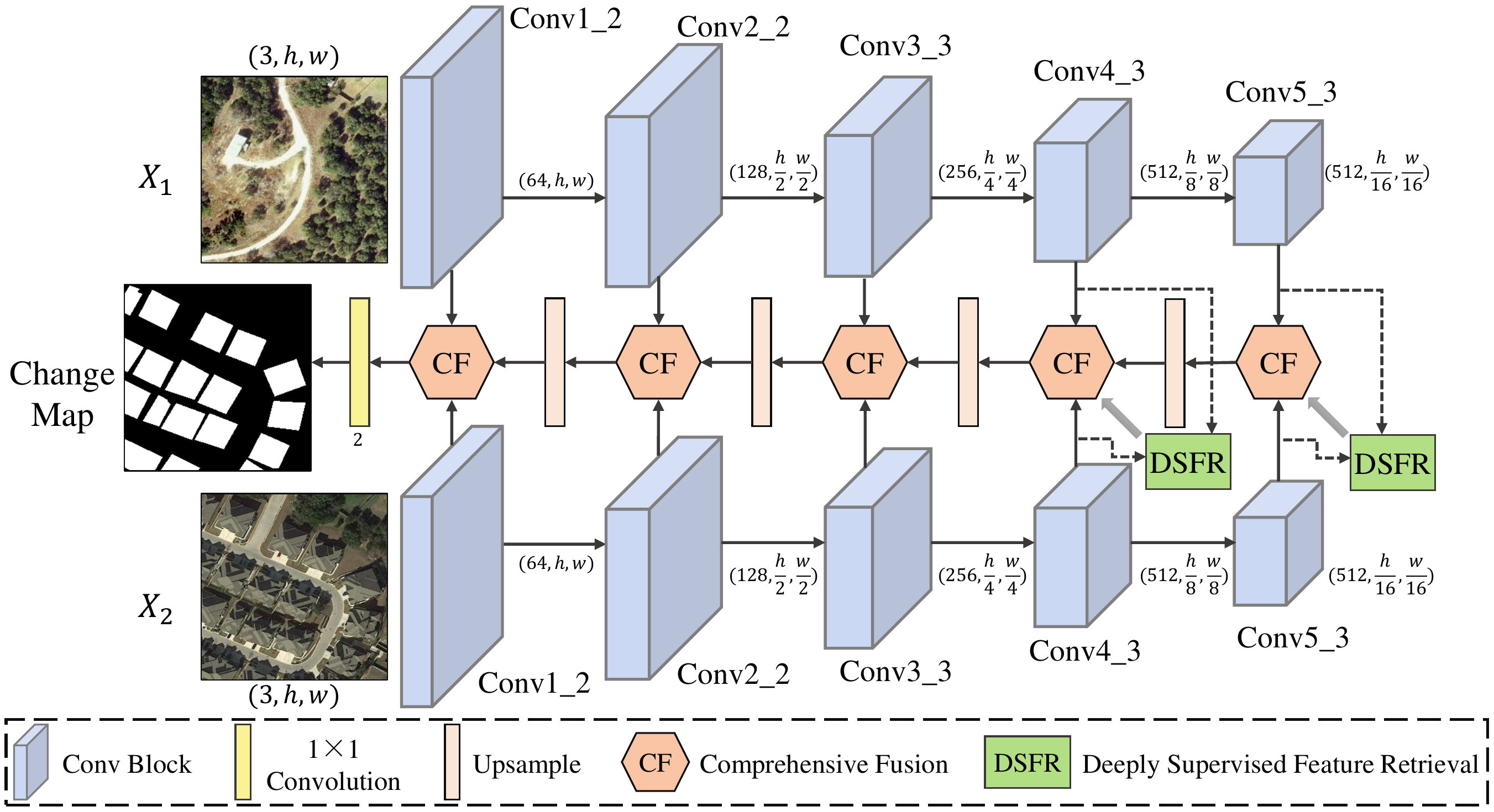}
    \caption{The overall architecture of the Dsfer-Net. The feature extractor is a Siamese architecture with shared weights. VGG-16 before pool5 is utilized as the backbone. DSFR modules are used for feature aggregation and retrieval in deep layers. CF blocks comprehensively fuse the multi-scale features.}
    \label{fig3}
\end{figure*}
\subsection{Modern Hopfield Networks}
Hopfield Networks, introduced in the 1980s \cite{hopfield1982neural, hopfield1984neurons}, are energy-based binary neural networks designed as associative memories with the specific purpose of storing and retrieving binary-valued data. Building on work in \cite{Chen:86,Psaltis:86,Baldi:87,Gardner:87,Abbott:87,krotov2016dense} proposing polynomial terms in the energy function, \cite{demircigil2017model} demonstrated that using exponential terms the storage capacity of these classical binary Hopfield networks can be considerably increased. Here, in contrast to these binary associative memory networks, we use modern Hopfield networks as introduced in \cite{Ramsauer:21} which work on continuous inputs, i.e. can store and retrieve real valued data. Crucially, unlike their classical counterparts, these modern Hopfield networks (MHNs) are differentiable and can thus be trained by gradient descent. They also retain the key ability of their binary counterparts to store exponentially many samples that can be retrieved in only one update step. Successful applications of MHNs include solving large-scale multi-instance learning tasks \cite{widrich2020modern}, predicting few- and zero-shot chemical reaction templates \cite{seidl2021modern} and conducting hierarchical prototype learning and improving text-image embedding representations for remote sensing image generation tasks \cite{xu2022txt2img}.

Modern Hopfield networks excel at associative memory tasks, making them also invaluable for associating changes in environmental features over time. The learning mechanisms of MHNs enable the deep learning network to continuously improve its ability to detect and classify changes based on feedback from observed data. To preserve deep feature representation and retrieval ability and to better understand and exploit the real changes from remote sensing data, we propose a deeply supervised feature retrieval module in this paper, building upon our previous study \cite{chang2022deep}. Additionally, in pursuit of explainable deep features learned by MHNs, we further explore appropriate applications of the prototype Hopfield layer for semantic segmentation.

\section{Methodology}
\subsection{Overview}
The overall architecture of the proposed Dsfer-Net is shown in Fig. \ref{fig3}. The paired images are input to a Siamese convolutional neural network to extract multi-level deep features. The encoder is built based on VGG-16 with pre-trained parameters whose final fully-connected layers and last pooling layer are removed. To decode the extracted multi-scale feature maps, we propose the comprehension fusion (CF) strategy and design deeply supervised feature retrieval (DSFR) modules. To capture semantic and sequential information from the deep layers, two DSFR modules are applied to aggregate and retrieve the feature difference in the hidden layers, and the multi-scale features from convolutional layers (Conv1$\_$2, Conv2$\_$2, Conv3$\_$3, Conv4$\_$3, Conv5$\_$3) are hierarchically integrated with retrieved features by the proposed CF strategy.

Let $X_1\in\mathbb{R}^{3\times h \times w}$ and $X_2\in\mathbb{R}^{3\times h \times w}$ denote the paired images captured from different times ($t_1$ and $t_2$), respectively, and $Y$ represents the annotated change map, where $h$ and $w$ represent the height and weight of the image, respectively. The main steps of this binary change detection task can be briefly described as follows:
\begin{figure}
    \centering
    \includegraphics[width=0.99\linewidth]{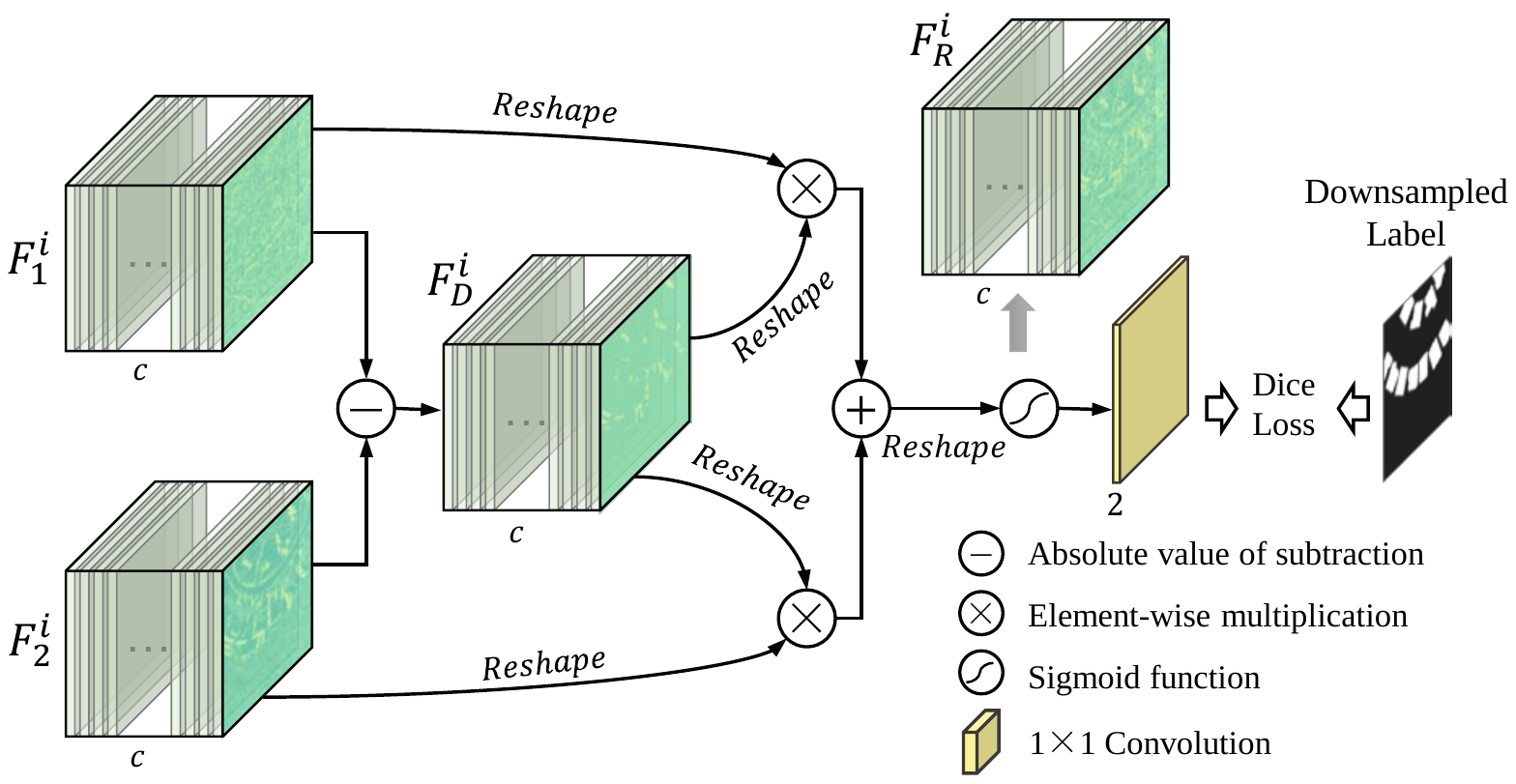}
    \caption{Deeply supervised feature retrieval (DSFR) module. $(F_1^i,F_2^i)$ represents the $i$-th paired feature maps extracted by the feature extractor. For the proposed network, the DSFR module is conducted on the fourth and fifth feature pairs, where $i= 4, 5$.}
    \label{fig4}
\end{figure}

\begin{itemize}
\item[1)] First, the image pair is fed into the feature extractor with shared weights, through which five pairs of multi-scale raw feature maps $\{(F_1^i, F_2^i)\}_{i=1}^5$ are obtained.
\item[2)] For higher-level features, i.e., $i=4,5$, the DSFR module is applied to obtain an intermediate change map that maintains the change of interests. The dice loss is calculated based on the change map and the downsampled label.
\item[3)] In the meantime, feature maps of two streams are concatenated and decoded by the CF strategy, and the change map $F_R^i$ retrieved from dual Hopfield layers of the DSFR module are forwarded, as the weight mask of concatenated features. Then the multi-scale features are merged and decoded in the same dimension, and the cross entropy loss is calculated based on the fused feature map and the ground-truth label.
\item[4)] Finally, the whole network is minimized by a combination of cross entropy loss and the dice losses using backpropagation.
\end{itemize}

\subsection{Deeply Supervised Feature Retrieval}
For the extracted deep feature maps, our intuition is that the changes between the feature pairs should also be preserved and traceable. However, the high-level features are usually too abstract and semantic to maintain the important texture information related to the change of interests, as has been previously shown in Fig. \ref{fig2}.

To exploit the semantic gaps in bitemporal image features and effectively aggregate sequential information, we introduce a deeply supervised feature retrieval (DSFR) module. Illustrated in Fig. \ref{fig4}, the DSFR module is primarily composed of dual Hopfield layers, strategically designed to exploit and consolidate sequential information from the input images. To accurately capture the changing information between features within deeper networks, the DSFR module adopts a supervision strategy and ultimately retrieves a map related to the change map. This architectural configuration enables our method to enhance the retrieval and synthesis of deep features, facilitating a more comprehensive understanding of temporal changes and semantic nuances within the image data. Given a deep feature pair $(F_1^i, F_2^i)$, the DSFR module calculates the absolute value of the difference feature, reshapes the features, and retrieves the difference feature through a Hopfield-type layer that treats the two respective deep features as the stored patterns. Then, the retrieved features are merged and normalized as a deep change map.

The prototype of the Hopfield layer, originally proposed in \cite{Ramsauer:21}, can learn or retrieve the state pattern in deep neural networks with multi-type structures and changeable application fields. To better fit MHNs to the proposed binary change detection task, the difference feature is defined as the state pattern to help activate the memory nodes that are highly associated with changing information stored in the raw features. Before passing to the Hopfield layer, the given features are first reshaped into 2-D matrices in which $F^i_D\in\mathbb{R}^{(h'w')\times c}$ denotes the 2-D absolute difference feature and the raw feature $F_j^i\in\mathbb{R}^{(h'w')\times c}, j=1,2$ denotes the store pattern, where $c$, $h'$, and $w'$ are the depth, height, and weight of the feature, respectively. For each Hopfield layer, the output of the retrieved difference feature is expressed as:
\begin{equation}
    \label{equ:mhnupdate}
    F^i_{R_j}=\text{softmax}(\beta F_D^{i}W_D W_{S}^\top F_j^{i\top})F_j^{i}W_{S},
\end{equation}
where $\text{softmax}(\cdot)$ denotes the row-wise softmax function, and $\beta$ is a scaling parameter. $W_D$, $W_{S}$ are learnable parameters that project the state pattern and store pattern, respectively, into a unified space. To further enhance the efficacy of the dual Hopfield layers, the retrieved features are merged and normalized as one change map:
\begin{equation}
    \label{equ:mhnupdate}
    F^i_{R}=\sigma(F^i_{R_1}+F^i_{R_2}),
\end{equation}
where $\sigma$ denotes a sigmoid function. 

The role of DSFR module has two aspects: 1) the retrieved deep change map directly flows forward to the CF block as a weight mask to highlight changed regions of the features extracted from shallower layers; and 2) the retrieved deep change map is fed into a $1\times 1$ convolutional layer to generate an intermediate change map $M_i\in\mathbb{R}^{2\times h'\times w'}$, and the dice loss is calculated according to the intermediate change map and the downsampled label.
\begin{figure}
    \centering
    \includegraphics[width=0.99\linewidth]{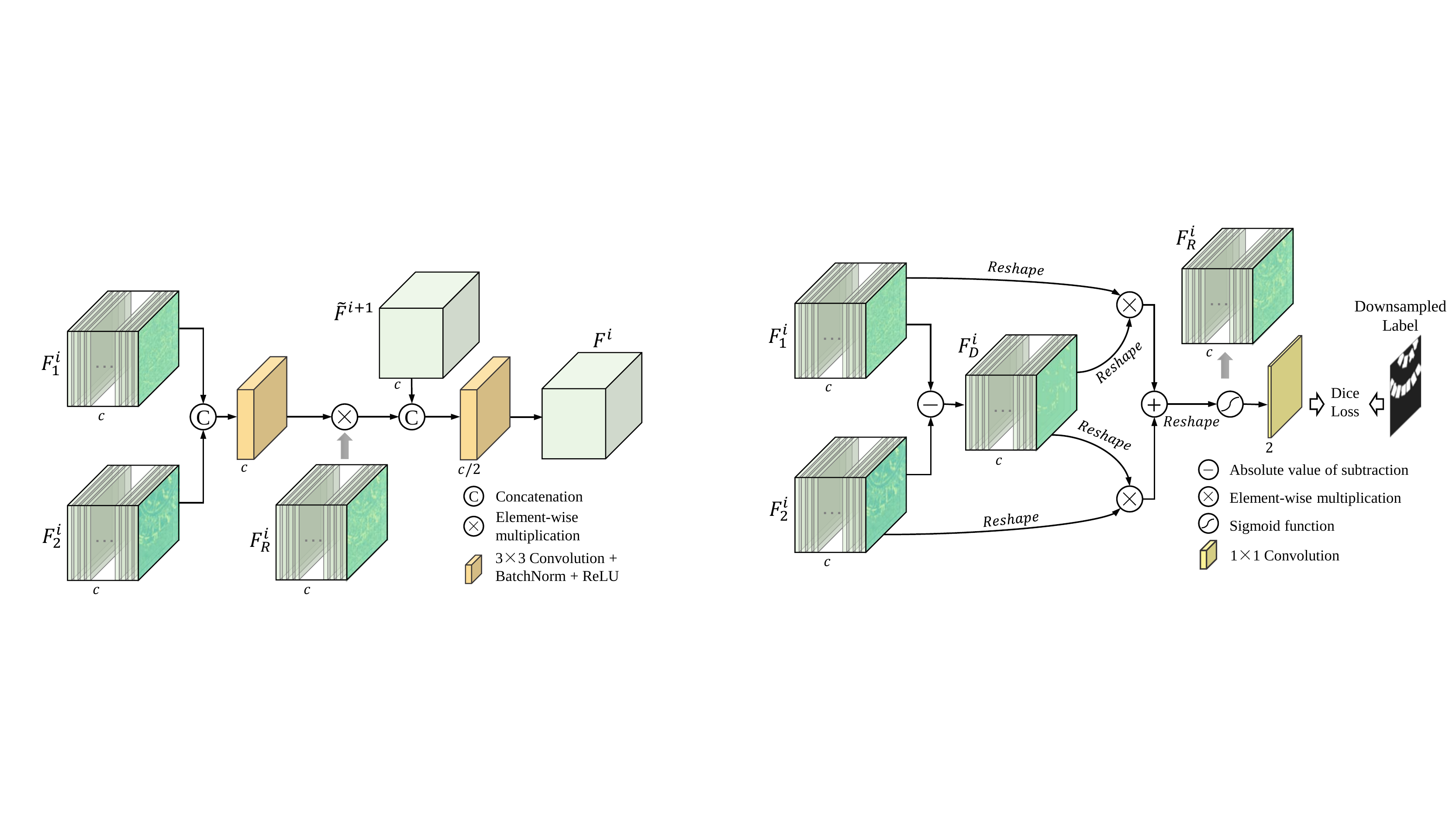}
    \caption{Comprehensive Fusion (CF) strategy. $(F_1^i,F_2^i)$ represents the $i$-th paired feature maps extracted by the feature extractor. If $i= 4, 5$, the concatenate features are multiplied with the retrieved feature $F_R^i$. $\Tilde{F}^{i+1}$ represents the upsampled feature obtained from the deeper CF block.}
    \label{fig5}
\end{figure}
\subsection{Comprehensive Fusion}
To fuse feature maps, the most straightforward approach involves directly concatenating the two stream features with those from deeper layers into a single feature map. However, the introduction of the DSFR module necessitates consideration of the retrieved feature reflecting changing weights. Directly combining heterogeneous features can lead to increased training difficulties. To address this challenge and more effectively incorporate the deep change map retrieved from the DSFR module into the raw feature maps, while also efficiently fusing multi-scale features, we devised a comprehensive fusion strategy termed the CF block, implemented within the decoder. For a detailed illustration of the CF strategy, please refer to Fig. \ref{fig5}. Given a raw feature pair $(F_1^i, F_2^i)$, the CF block first concatenates these two features and feeds them into a $3\times 3$ convolutional layer. For the deepest layers, i.e., $i=4$ and $5$, the retrieved change map $F_R^i$ is input as a mask of the concatenated features. The multiplied feature emphasizes the changed information and suppresses the unchanged regions. The multiplied feature then concatenates the upsampled CF feature $\Tilde{F}^{i+1}$ from deeper layers and decodes the concatenated feature through another $3\times 3$ convolutional layer as the output map of the $i$-th CF block of the whole network. Formally, the CF feature map of the $i$-th stage can be expressed as
\begin{equation}
    F^i=\mathnormal{f}_2^{(3\times 3)}( \textbf{Concat}(\mathnormal{f}_1^{(3\times 3)}(\mathcal{F}^i)\odot F_R^i,\Tilde{F}^{i+1})),
\end{equation}
where each $f^{(3\times 3)}(\cdot)$ denotes a convolution layer with kernel size $3\times 3$ and the appropriate number of channels linked to $i$ as shown in Fig. 5, followed by a batch normalization layer (BN) and a ReLU activation layer, $\textbf{Concat}(\cdot, \cdot)$ denotes the concatenation of two features, $\mathcal{F}^i=\textbf{Concat}(F_1^i,F_2^i)$ represents the concatenated feature of $F_1^i$ and $F_2^i$, and $\odot$ denotes element-wise multiplication.

The CF block operates sequentially, enhancing change information by integrating the retrieved change map with raw feature maps, followed by fusion with features from deeper layers. This comprehensive fusion strategy, compared to direct concatenation, effectively leverages change-reflected information while integrating multiple feature maps from shallower and deeper layers.
\subsection{Network Training}

It has been proved that the binary cross-entropy loss function performs effectively for two-class problems by treating the softmax layer as a pixel-wise classification problem \cite{lei2020hierarchical}. However, for typical change detection datasets, the class distribution of changed and unchanged pixels is usually heavily imbalanced. To relieve this class-imbalance problem, we adopt the weighted binary cross-entropy loss function to train our model. This function can be expressed as follows

\begin{equation}
    \mathcal{L}_{wbce}=-\frac{1}{hw}\sum_{l=1}^{hw}(w_1 Y_l\log\hat{Y}_l+w_0 (1-Y_l)\log(1-\hat{Y}_l)),
\end{equation}
where $w_0$ and $w_1$ represent the weight for the unchanged class and the changed class, respectively, and $Y_l$ and $\hat{Y}_l$ denote the $l$-th pixel of the annotated change map and the predicted change map, respectively.

In the DSFR module, the dice loss is also calculated for deeper convolutional features. Let $i$ denote the $i$-th basic stage of the backbone network; thus, the downsampled annotated change map and the intermediate predicted change map can be respectively represented as $Y^i$ and $\hat{Y}^i$, and the dice loss can be written as
\begin{equation}
    \mathcal{L}^i_{dice}=1-2\cdot\frac{\sum_{m=1}^{h'w'}Y^i_m\hat{Y}^i_m}{\sum_{m=1}^{h'w'}(Y^i_m+\hat{Y}^i_m)},
\end{equation}
where $Y^i_m$ and $\hat{Y}^i_m$ denote the $m$-th pixel of the downsampled label and the intermediate prediction, respectively. In this study, we can derive the relationships of the heights and weights between the original annotated change map and the intermediate map are
\begin{equation}
    \begin{split}
        &h' = 2^{1-i}\cdot h, \\
        &w' = 2^{1-i}\cdot w.
    \end{split}
\end{equation}

Finally, the objective function for training the proposed network can be formulated as:

\begin{equation}
    \mathcal{L}_{total}=\mathcal{L}_{wbce}+\lambda\cdot(\mathcal{L}_{dice}^4 +\mathcal{L}_{dice}^5)/2,
\end{equation}
where $\lambda$ is the weighting factor to adjust respective loss terms. 
\section{Experiments and Discussion}
In this study, the proposed Dsfer-Net is conducted on three widely used CD datasets and compared with eight state-of-the-art algorithms. Detailed implementation and evaluation metrics are illustrated bellow. Finally, the experimental results are provided and discussed in detail.
\subsection{Dataset Description and Evaluation Metrics}
We carried out the experiments on three CD datasets:

1) \textbf{LEVIR-CD} is a large-scale building change detection dataset that covers 20 different regions in Texas, US. It contains 637 very high-resolution (VHR) Google Earth (GE) image patch pairs with a size of 1024$\times$1024 pixels and a spatial resolution of 0.5 m/pixel. Following \cite{chen2020spatial}, we divided the image pairs into training, validation, and test datasets with an approximate ratio of 7:1:2. The original images were cropped into 16 small patches of sizes of 256$\times$256. Thus, there is a total of 7,120 image pairs for training, 1,024 for validation, and 2,048 for testing.

2) \textbf{WHU-CD} is a subset of the WHU Building dataset \cite{ji2018fully} which contains both aerial and satellite imagery. The building change subset covers an area of 20.5 km$^2$ in Christchurch, New Zealand where a 6.3-magnitude earthquake occurred in February 2011 and was rebuilt in the following years. One image pair with a size of 32,507$\times$15,354 pixels and a spatial resolution of 0.075 m/pixel was collected in 2012 and 2016, respectively. To conduct the experiments, we cropped the original images into 8189 non-overlapped tiles with 256$\times$256 pixels and split the cropped tiles into a training set, a validation set, and a test set with a ratio of 6:1:3.  Thus, there is a total of 4,572 image pairs for training, 762 for validation, and 2,286 for testing.

3) \textbf{CDD} \cite{lebedev2018change} is a season-varying remote sensing dataset collected from Google Earth (DigitalGlobe) that consists of seven pairs of manually ground-truth created season-varying images with 4725$\times$2700 pixels and four pairs of season-varying images with minimal changes and resolution of 1900$\times$1000 pixels. The spatial resolution of the images is 0.03$\sim$1 m/pixel. By cropping the original images with a size of 256$\times$256 and a randomly rotated fragment (0$\sim$2$\pi$), 16,000 pairs were obtained, with 10,000 and 3,000 pairs used for training and validation, respectively, and the remaining 3,000 pairs were used for testing.

To quantitatively evaluate the performance of the proposed method, precision (P), recall (R), F1-score, overall accuracy (OA), and Intersection over Union (IoU) were chosen as the evaluation metrics, which are defined as follows:
\begin{equation*}
    \begin{split}
        &\text{P}=\frac{\text{TP}}{\text{TP}+\text{FP}} \\
        &\text{R}=\frac{\text{TP}}{\text{TP}+\text{FN}} \\
        &\text{F1}=\frac{2\text{P}\text{R}}{\text{P}+\text{R}} \\
         &\text{OA}=\frac{\text{TP}+\text{TN}}{\text{TP}+\text{FP}+\text{TN}+\text{FN}} \\
        &\text{IoU}=\frac{\text{TP}}{\text{TP}+\text{FP}+\text{FN}} \\
    \end{split}
\end{equation*}
where TP, TN, FP, and FN represent the number of true positives, true negatives, false positives, and false negatives, respectively.

\subsection{Comparative Methods and Implementation Settings}
To validate the performance of our model, we compared the proposed Dsfer-Net with eight state-of-the-art approaches. A brief introduction to the comparative methods follows.
\begin{itemize}
    \item FC-Siam-conc \cite{daudt2018fully} is a U-Net-based architecture that applies a Siamese encoding stream in order to extract deep features from both images separately; then it concatenates the extracted features with two skip connections and passes them to the decoding stream.
    \item FC-EF-Res \cite{daudt2019multitask} is an early fusion architecture that first concatenates the bitemporal images together and passes them through the network. Unlike traditional convolutional layers, residual blocks have been added to improve network performance and facilitate training.
    \item STANet \cite{chen2020spatial} designs a self-attention mechanism to model the spatial-temporal relationships of the bitemporal images and capture the dependencies at various scales. The attentive weights between the features are then calculated to generate more discriminative features.
    \item BiT \cite{chen2021remote} is a transformer-based module that uses visual words to express the space-time changes from high-level concepts. The Bitemporal image Transformer (BiT) is incorporated in a deep feature differencing-based network for change detection.
    \item DSAMNet \cite{shi2021deeply} employs two deeply supervised layers to assist the learning of the hidden layer and designs a metric module to stack the resized features of different scales. Convolutional block attention modules (CBAMs) are adopted to make the features more discriminative.
    \item ChangeFormer \cite{bandara2022transformer} is a transformer-based network that integrates a hierarchical transformer encoder with a Multi-Layer Perception (MLP) decoder in a Siamese network architecture. This design efficiently captures multi-scale long-range details.
    \item VcT \cite{jiang2023vct} employs the Visual Change Transformer as its backbone, effectively leveraging both intra-image and inter-image cues by capturing dependencies between reliable tokens in dual images.
    \item FrNet \cite{chang2022deep} is a preliminary study that utilizes the MHNs to integrate deep features. The Hopfield pooling layer is utilized together with the deepest convolutional layer, and the VGG-16 is selected as the feature extractor.
\end{itemize}

All the experiments were implemented on Pytorch and trained using an NVIDIA A100 GPU. To ensure a fair comparison of the efficiency of the comparative methods, we considered both the parametric settings specified in the original literature and the actual performance, reporting the best performance. For the proposed Dsfer-Net, we set the projection dimension of the Hopfield layers to 512. We adopted the Adam optimizer with an initial learning rate of  $1\text{e}-5$ for the LEVIR-CD and the WHU-CD datasets, and $1\text{e}-4$ for the CDD dataset. The learning rate will linearly decay to 0 after 30,000 iterations. We set the weight decay to $1\text{e}-5$ and the batch size to 32. Validation is performed after every 500 iterations, and the best model on the validation set is used for evaluation on the test set. The same training, validation, and testing settings are adopted for all three datasets.

\subsection{Benchmark Comparison and Analysis}
\begin{figure*}
    \centering
    \includegraphics[width=\linewidth]{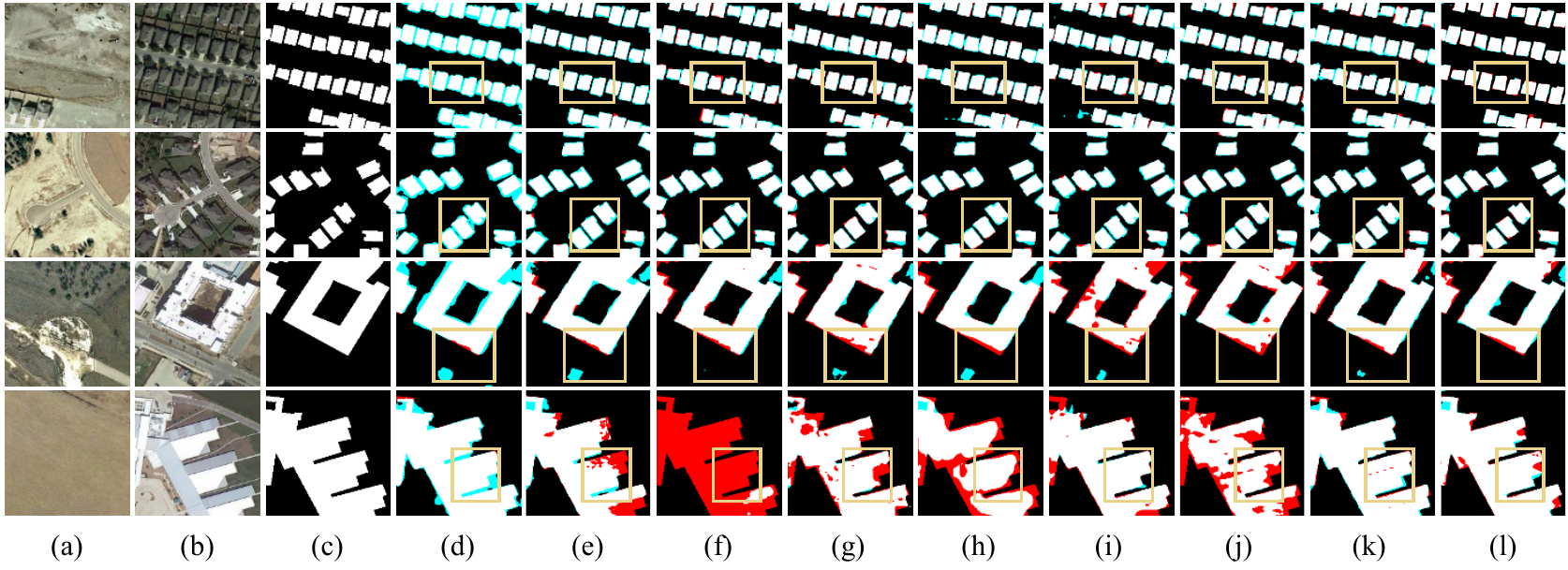}
    \caption{Visualization examples of different methods on the LEVIR-CD dataset. (a) Image from $t_1$. (b) Image from $t_2$. (c) Ground-truth. (d) FC-Siam-conc. (e) FC-EF-Res. (f) STANet. (g) BiT. (h) DSAMNet. (i) ChangeFormer. (j) VcT. (k) FrNet. (l) Dsfer-Net. TP, FP, TN, and FN are represented in white, cyan, black, and red, respectively.}
    \label{fig6}
\end{figure*}
\begin{figure*}
    \centering
    \includegraphics[width=\linewidth]{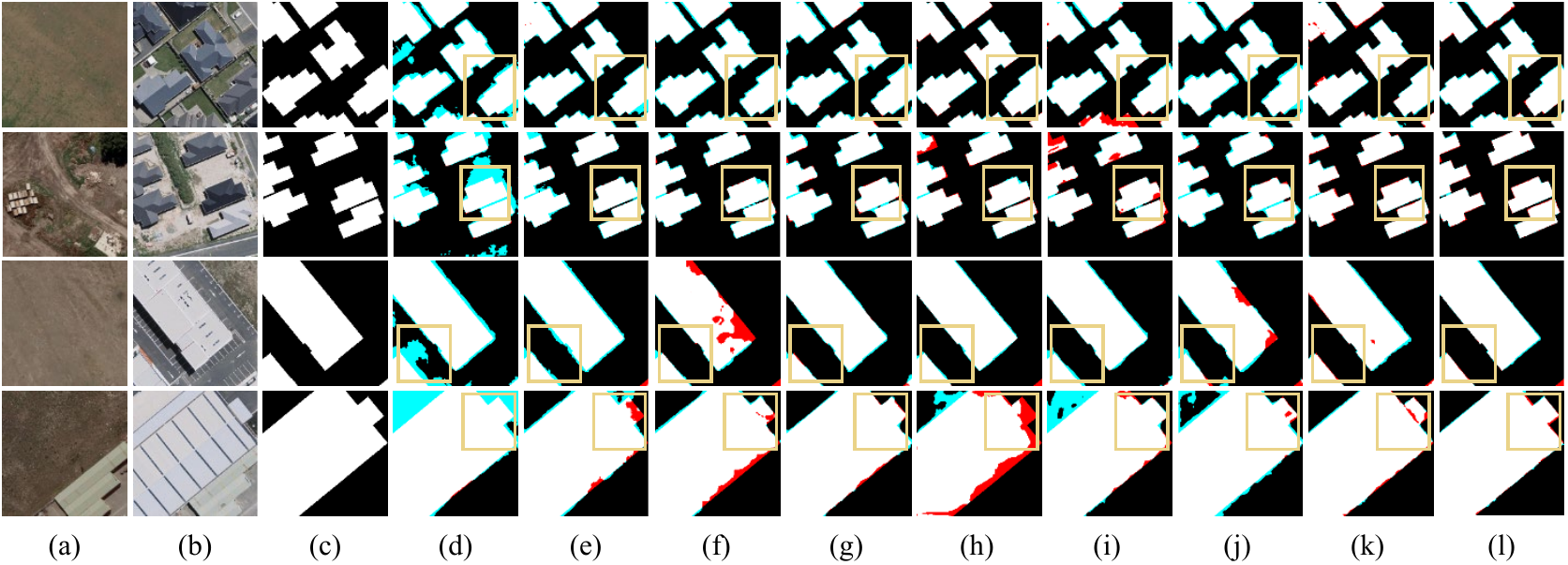}
    \caption{Visualization examples of different methods on the WHU-CD dataset. (a) Image from $t_1$. (b) Image from $t_2$. (c) Ground-truth. (d) FC-Siam-conc. (e) FC-EF-Res. (f) STANet. (g) BiT. (h) DSAMNet. (i) ChangeFormer. (j) VcT. (k) FrNet. (l) Dsfer-Net. TP, FP, TN, and FN are represented in white, cyan, black, and red, respectively.}
    \label{fig7}
\end{figure*}
\begin{figure*}
    \centering
    \includegraphics[width=\linewidth]{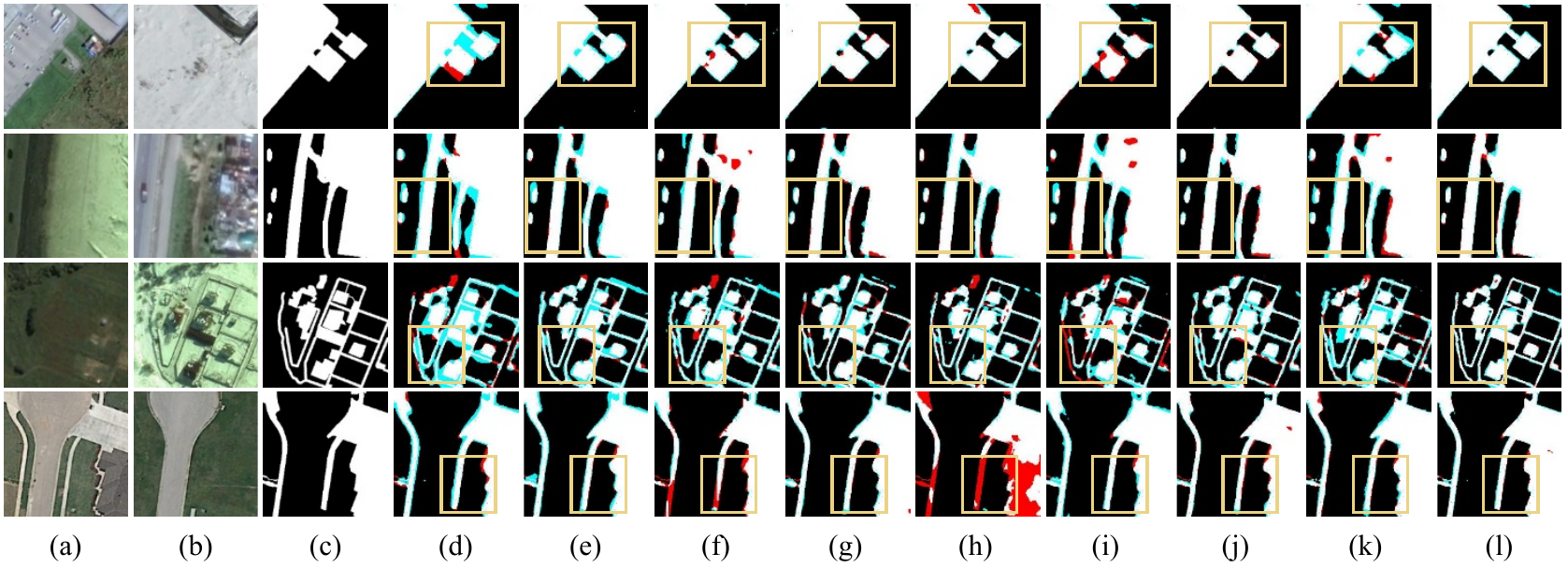}
    \caption{Visualization examples of different methods on the CDD dataset. (a) Image from $t_1$. (b) Image from $t_2$. (c) Ground-truth. (d) FC-Siam-conc. (e) FC-EF-Res. (f) STANet. (g) BiT. (h) DSAMNet. (i) ChangeFormer. (j) VcT. (k) FrNet. (l) Dsfer-Net. TP, FP, TN, and FN are represented in white, cyan, black, and red, respectively.}
    \label{fig8}
\end{figure*}
In this section, the proposed model is compared with eight state-of-the-art methods on three benchmark datasets. The visualized results are shown in Figs. \ref{fig6} to \ref{fig8}. In the LEVIR-CD and WHU-CD datasets, buildings are the changed objects to be evaluated. It can be observed that most of the methods can correctly detect small buildings in both the LEVIR-CD and WHU-CD datasets. However, FC-Siam-conc and FC-EF-Res failed to distinguish the unchanged landcovers surrounding the buildings, resulting in a high incidence of false alarms compared to other methods. This indicates that directly concatenating deep features or the feature differences cannot effectively provide discriminative information for network training. For large-size buildings, as highlighted by the square, STANet, BiT, DSAMNet, and VcT cannot correctly detect the objects, especially those with rather complex shapes. The results show that although the well-received attention mechanism is capable of reintegrating and emphasizing interesting information about deep features, its application reveals that it suffers from a self-retrieval mechanism and may fail to highlight real differences between the temporal images. Compared to FrNet, the proposed model can better control false alarms and provide clearer boundaries between changed objects. Among all methods, ours performs the best in detecting changed objects and returning the most accurate contours of buildings. We conclude that the designed DSFR module is helpful in extracting real changes from time-series data and preserving details as close to the ground truth as possible.

The CDD dataset records seasonal variations and object changes on the ground at different times. The proposed method achieves better results than other comparison methods. As illustrated in Fig. \ref{fig8}, buildings, roads, and cars have been chosen as the objects of change for visualization, reflecting changes influenced by time, extreme weather, and human activities. Visualized results show that our model can extract both small-sized and large-sized changes in the CDD dataset, and has the fewest false positives (FP) and false negatives (FN) compared to other methods. Even in relatively complex semantic scenes, the proposed model can accurately detect real changes. In contrast to our model, FC-Siam-conc and FC-EF-Res cannot avoid false alarms when detecting changes. Furthermore, they perform poorly in preserving the boundaries of narrow roads and small objects. Although STANet, BiT, ChangeFormer, and FrNet can extract changing objects that have relatively large areas, they still perform poorly in detecting fine-grained changes. The DSAMNet, as a deeply supervised and attention-based method, performs well in identifying narrow and small changes, but suffers from missed detection of large-sized objects and fails to retain the internal compactness of objects. Among all comparable methods, VcT, as a transformer-based method, has balanced detection performance in identifying narrow objects and relatively large changed areas.
\begin{table*}
\centering
\caption{Quantitative Comparison Results on Three Datasets.}
\begin{threeparttable}
\begin{tabular}{c|ccccc|ccccc|ccccc}
\toprule 
\multirow{2}{*}{\textbf{Method}} & \multicolumn{5}{c|}{\textbf{LEVIR-CD}} & \multicolumn{5}{c|}{\textbf{WHU-CD}} & \multicolumn{5}{c}{\textbf{CDD}} \\ 
                  & P   & R   & F1  & OA & IoU  & P   & R   & F1   & OA  & IoU  & P   & R   & F1  & OA   & IoU   \\ \midrule
FC-Siam-conc      & 68.29   & \textbf{96.93}   & 80.12 &97.55 &66.84  & 64.95 &\textbf{93.57} &  76.68 &  97.90 & 62.18 &77.31 &88.94 & 82.72 & 95.61 & 70.53 \\ 
FC-EF-Res         & 79.71   & 93.62   & 86.11 &98.46& 75.61& 87.92 & 86.49 & 87.20 &  99.06 & 77.31& 85.66& 94.85 & 90.02 & 97.52 &  81.86 \\ 
STANet            & 86.55   & 87.99    & 87.27& 98.69& 77.41 & 90.55 & 83.75&87.02  &  99.08 & 77.02  &88.92 & 90.83& 89.86&97.58 & 81.59 \\
BiT               & 86.30    & 90.17    & 88.19 & 98.77& 78.87 & 89.57 & 85.72 & 87.60 & 99.10 & 77.94 & 92.49& 91.22 & 91.85 &98.09 & 84.92 \\ 
DSAMNet       & 90.00 & 82.49 & 86.08& 98.52  & 75.56&83.49 & 90.18& 86.71& 99.05 &    76.53 & 93.27 & 92.54 &92.91 & 98.32 & 86.76 \\ 
ChangeFormer &81.36 &87.77 &84.45 &98.35 &73.08 &85.01 &79.61 & 82.22 & 98.73 & 69.80 &85.12 &90.07 &87.53 &96.97 &77.82  \\
VcT &85.67 &88.14 &86.89 &98.64 &76.81 &84.04 &88.84 &86.37 &98.96 & 76.02 &93.19 &89.38 & 91.25 & 97.98 & 83.90\\
FrNet           & 86.80  & 90.91  & 88.81& 98.83  & 79.87& 93.76 & 88.72& 91.17& 99.37 &    83.77 & 89.83 & 91.52 &90.67 & 97.78 & 82.92 \\ \midrule
Dsfer-Net         & \textbf{89.42}    & 92.15  &  \textbf{90.76} & \textbf{99.04}  &\textbf{83.09}  &  \textbf{94.17}    & 91.04 & \textbf{92.58} & \textbf{99.46}  & \textbf{86.18}  & \textbf{93.70}  & \textbf{96.69}  &\textbf{95.17} & \textbf{98.84}  & \textbf{90.79}  \\ \bottomrule
\end{tabular}
\label{tab1}
\begin{tablenotes}
\footnotesize
\item[*] Note: All the scores are reported in percentage ($\%$), and the best results are highlighted in \textbf{Bold}.
\end{tablenotes}
\end{threeparttable}
\end{table*}

The quantitative evaluations of all methods are illustrated in Table \ref{tab1}. Given that our method is proposed for binary change detection, we focus solely on reporting the accuracy of the changed class. Across all three datasets, our proposed model exhibits superior results in Precision, F1 score, OA, and IoU. In the LEVIR-CD dataset, the F1 score and IoU of Dsfer-Net exceed those of the best-performing comparative methods by 1.9 and 3.2 points, respectively. In the WHU-CD dataset, our model outperforms FrNet by 1.4 points in F1 score and 2.4 points in IoU. In the CDD dataset, the proposed method achieves a 2.2 points higher F1 score and a 3.9 points higher IoU than DSAMNet. Additionally, we observe a negative correlation between precision and recall in the methods, which can be intuitively understood as false alarms and missed detections usually not occurring simultaneously. Therefore, change detection methods need to strike a balance between detecting more positives and controlling false alarms. In the quantitative comparisons across various datasets, our proposed method demonstrates superior performance with fewer false alarms and reasonable missed detections.
\begin{table}
\centering
\caption{Parameters and Inference Time (in Seconds) on the LEVIR-CD dataset.}
\begin{tabular}{c|cc}
\toprule 
\textbf{Methods} & \textbf{Params(M)}  & \textbf{Inference Time(S)}  \\ \midrule
FC-Siam-conc      & 1.55M    & 16.60  \\ 
FC-EF-Res         & 1.10M    & 18.68  \\ 
STANet            & 16.93M   & 40.33  \\
BiT               & 11.94M   & 39.60  \\ 
DSAMNet           & 16.95M   & 27.91  \\ 
ChangeFormer      & 43.99M   & 107.27 \\
VcT               & 12.41M   & 82.23  \\
FrNet             & 15.00M   & 15.22  \\ \midrule
Dsfer-Net         & 34.08M   & 21.27  \\ \bottomrule
\end{tabular}
\label{tab2}
\end{table}

To better understand the efficiency of our model, we provide a comparison of model parameters (Params) and inference time between our model and other comparative methods on the LEVIR-CD dataset. All experiments were conducted on a server equipped with an NVIDIA A100 GPU. As depicted in Table \ref{tab2}, the parameter count of our proposed Dsfer-Net model is 34.08M. This is primarily due to the presence of the Hopfield layer, which requires a larger computational memory. However, despite the higher parameter count, our proposed method demonstrates competitive inference times. This performance is comparable to that of attention-based methods.

\subsection{Ablation Study}
\begin{table}[]
\centering
\caption{Performance Contribution of Different Strategies in Dsfer-Net.}
\begin{threeparttable}
\begin{tabular}{c|l|ccc}
\toprule
                 & \textbf{CF} \textbf{DSFR} & \textbf{LEVIR-CD} & \textbf{WHU-CD} &\textbf{CDD}   \\\midrule
Base+Concat   &                           &   89.97     &91.87           &93.91 \\
Base+CF       & \ \checkmark              &   90.05     &91.88           &94.70 \\
Base+Concat+DSFR& \ \ \ \ \ \ \ \checkmark  &   90.28     &92.34          &94.33 \\
Base+CF+DSFR  & \ \checkmark  \ \ \ \  \checkmark & \textbf{90.76} & \textbf{92.58}  & \textbf{95.17} \\\bottomrule                
\end{tabular}
\label{tab3}
\begin{tablenotes}
\footnotesize
\item[*] Note: All the scores are reported in percentage ($\%$), and the best F1 scores are highlighted in \textbf{Bold}.
\end{tablenotes}
\end{threeparttable}
\end{table}

To evaluate the effectiveness of our designed CF strategy and DSFR module, we conducted ablation studies on three datasets: Base+Concat, Base+CF, Base+Concat+DSFR, and Base+CF+DSFR. As the name suggests, Base+Concat is a Siamese network with VGG-16 as the backbone, using concatenated features as input for the decoder. Base+CF integrates multi-scale features using the CF strategy. In Base+Concat+DSFR, the DSFR module retrieves changed information from deeper layers, with features from each convolutional block fused through concatenation. Base+CF+DSFR corresponds to the proposed Dsfer-Net. The F1 scores of the different configurations are illustrated in Table \ref{tab3}. It is observed that the CF strategy brings slight improvements over the classic feature concatenation process across all three datasets. Additionally, combining the CF strategy with the DSFR module enhances change detection performance further compared to feature concatenation alone.

\begin{table}
\centering
\caption{Parameters and Inference Time (in Seconds) on the LEVIR-CD dataset.}
\begin{tabular}{c|cccc}
\toprule 
\textbf{Backbone} & \textbf{F1}&\textbf{OA}&\textbf{Params(M)}  & \textbf{Inference Time(S)}  \\ \midrule 
ResNet-18       & 88.24 & 98.75 & 22.90M  & 18.12 \\ 
Transformer     & 86.98 & 98.64 & 59.24M  & 57.08 \\
VGG-16          & 90.76 & 99.04 & 34.08M  & 21.27  \\ \bottomrule
\end{tabular}
\label{tab21}
\end{table}

To further evaluate the effectiveness of our approach, we conducted benchmark experiments on the LEVIR-CD dataset using different backbone architectures with our Dsfer-Net model. Table \ref{tab21} presents the results obtained by employing three classical backbones, ResNet-18, Transformer, and VGG-16, to extract bi-temporal deep features. To maintain consistency, we preserved the parametric settings from our previous experiments conducted with the VGG-16 architecture.

Our findings reveal that Dsfer-Net achieves optimal performance when leveraging the VGG-16 architecture, outperforming both ResNet-18 and transformer encoders while maintaining a balanced parameter count and inference time. Notably, when comparing our model with other methods listed in Table \ref{tab1}, particularly those employing ResNet-18 as the backbone, such as STANet, BiT, and DSAMNet, Dsfer-Net demonstrates superior performance on the LEVIR-CD dataset. Furthermore, in comparison to ChangeFormer and VcT, which utilize transformer-based encoders, our model equipped with a transformer-based encoder also exhibits better performance. These results underscore the efficacy of our approach across different backbone architectures and highlight its competitive advantage in change detection tasks.

\subsection{Sensitivity Analysis}
\begin{table}[]
\centering
\caption{Sensitivity Analysis of Weighting Factor on Loss Function.}
\begin{threeparttable}
\begin{tabular}{c|cc|cc|cc}
\toprule
\multirow{2}{*}{$\lambda$} & \multicolumn{2}{c|}{\textbf{LEVIR-CD}}   & \multicolumn{2}{c|}{\textbf{WHU-CD}}  & \multicolumn{2}{c}{\textbf{CDD}}                        \\
         & \multicolumn{1}{c}{F1} & \multicolumn{1}{c|}{OA} & \multicolumn{1}{c}{F1} & \multicolumn{1}{c|}{OA} & \multicolumn{1}{c}{F1} & \multicolumn{1}{c}{OA} \\ \midrule
0.001    & 90.69     & 99.04       & 92.58    & 99.46      & 95.01       & 98.80                    \\
0.01     & 90.73     & 99.04       & \textbf{92.58}    & \textbf{99.46}    & 94.83     & 98.75      \\
0.1      & \textbf{90.76} & \textbf{99.04}  & 92.56    & 99.46      & 94.99     & 98.79                   \\
1        & 90.66     & 99.03       & 92.34    & 99.45      & \textbf{95.17}    & \textbf{98.84}           \\
10       & 90.12     & 98.96       & 92.04    & 99.43      & 94.75       & 98.75     \\
100      & 89.47     & 98.88       & 92.75    & 99.48      & 94.49       & 98.68     \\\bottomrule
\end{tabular}
\label{tab4}
\begin{tablenotes}
\footnotesize
\item[*] Note: All the scores are reported in percentage ($\%$), and the best results are highlighted in \textbf{Bold}.
\end{tablenotes}
\end{threeparttable}
\end{table}
As described in Section III, the loss function of the proposed network is constructed by the cross entropy loss and the dice losses balanced by a weighting factor. To explore the influence of different values on training the proposed network, comparative experiments are conducted on three datasets by setting different  $\lambda$ values. Since the DSFR modules are adopted to improve feature extraction capability and preserve texture information in the deep layers, the best value of $\lambda$ represents the most appropriate proportion of the retrieval features for the training process. Note that when $\lambda$ is set to 0, the network is equivalent to the second baseline “Base+CF” in the previous subsection. As can be seen in Table \ref{tab4}, a relatively smaller value of $\lambda$ is more beneficial to the performance. It is known that the shape of changed objects will slightly change and texture information of details will also be lost after repeated downsampling. Adding appropriate weights to the deep supervision modules can help capture the location of changed objects across scenes, but higher weights may cause false alarms on fine-grained landcovers and result in poorer performances. Specifically, for the LEVIR-CD, WHU-CD, and CDD datasets, the best performance is obtained when the values of $\lambda$ equal to $0.1$, $0.01$, and 1, respectively. 

\subsection{Visualizations of the Hopfield Features}
To further evaluate the effectiveness of the DSFR modules, we provide visualizations of the retrieved features across three datasets. The DSFR module primarily consists of a dual Hopfield layer and is implemented in the fourth and fifth convolutional stages. Thus, the aggregated Hopfield features are compared with the downsampled image pairs and labels of the same size.

\begin{figure}
    \centering
    \includegraphics[width=0.99\linewidth]{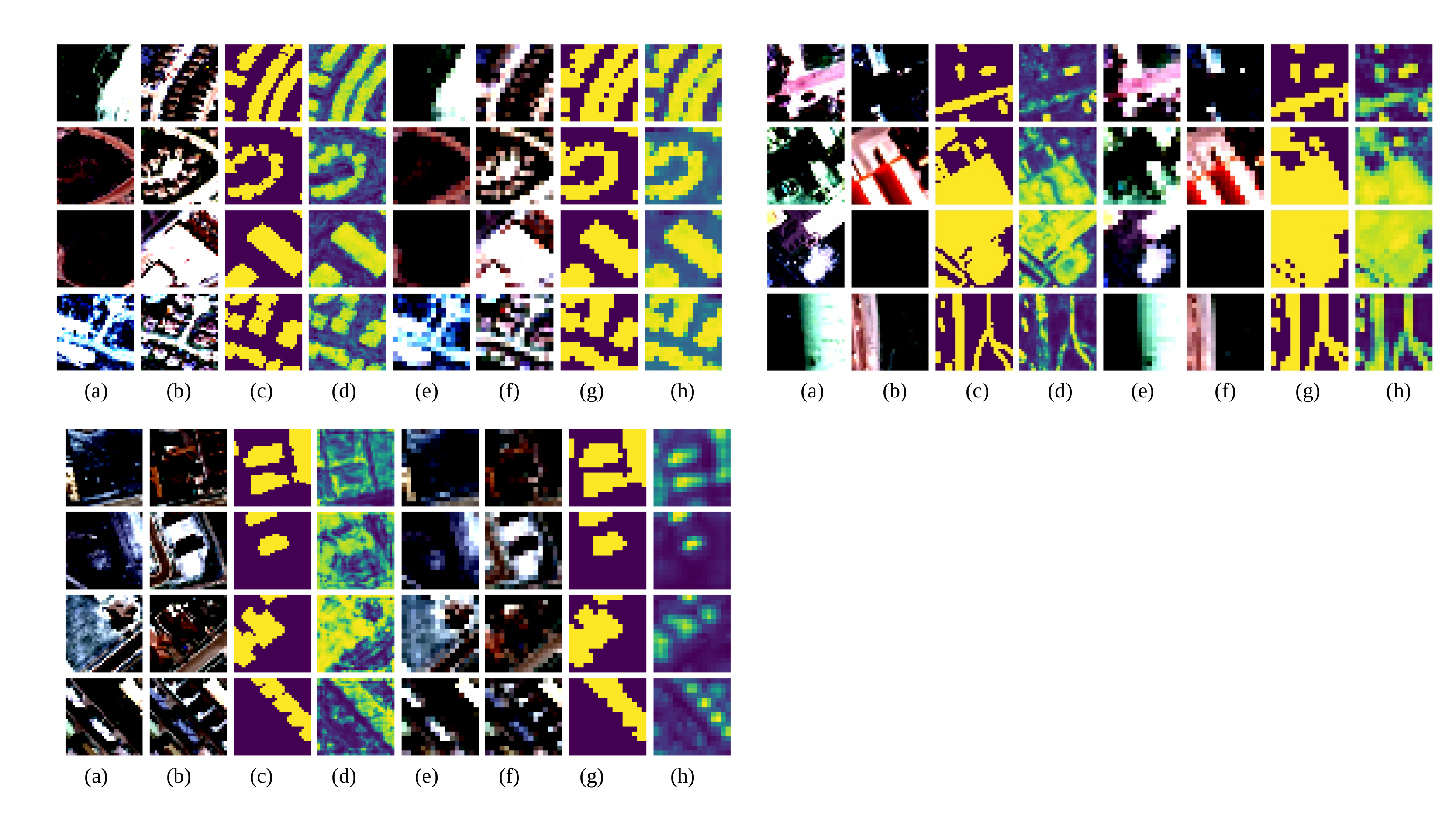}
    \caption{Visualization examples of deep features retrieved by DSFR modules in the LEVIR-CD dataset. (a) Images from $t_1$  downsampled by 8. (b) Images from $t_2$ downsampled by 8. (c) Ground-truths  downsampled by 8. (d) Retrieved features from the fourth stage. (e) Images from $t_1$  downsampled by 16. (f) Images from $t_2$ downsampled by 16. (g) Ground-truths downsampled by 16. (h) Retrieved features from the fifth stage.}
    \label{fig9}
\end{figure}

\begin{figure}
    \centering
    \includegraphics[width=0.99\linewidth]{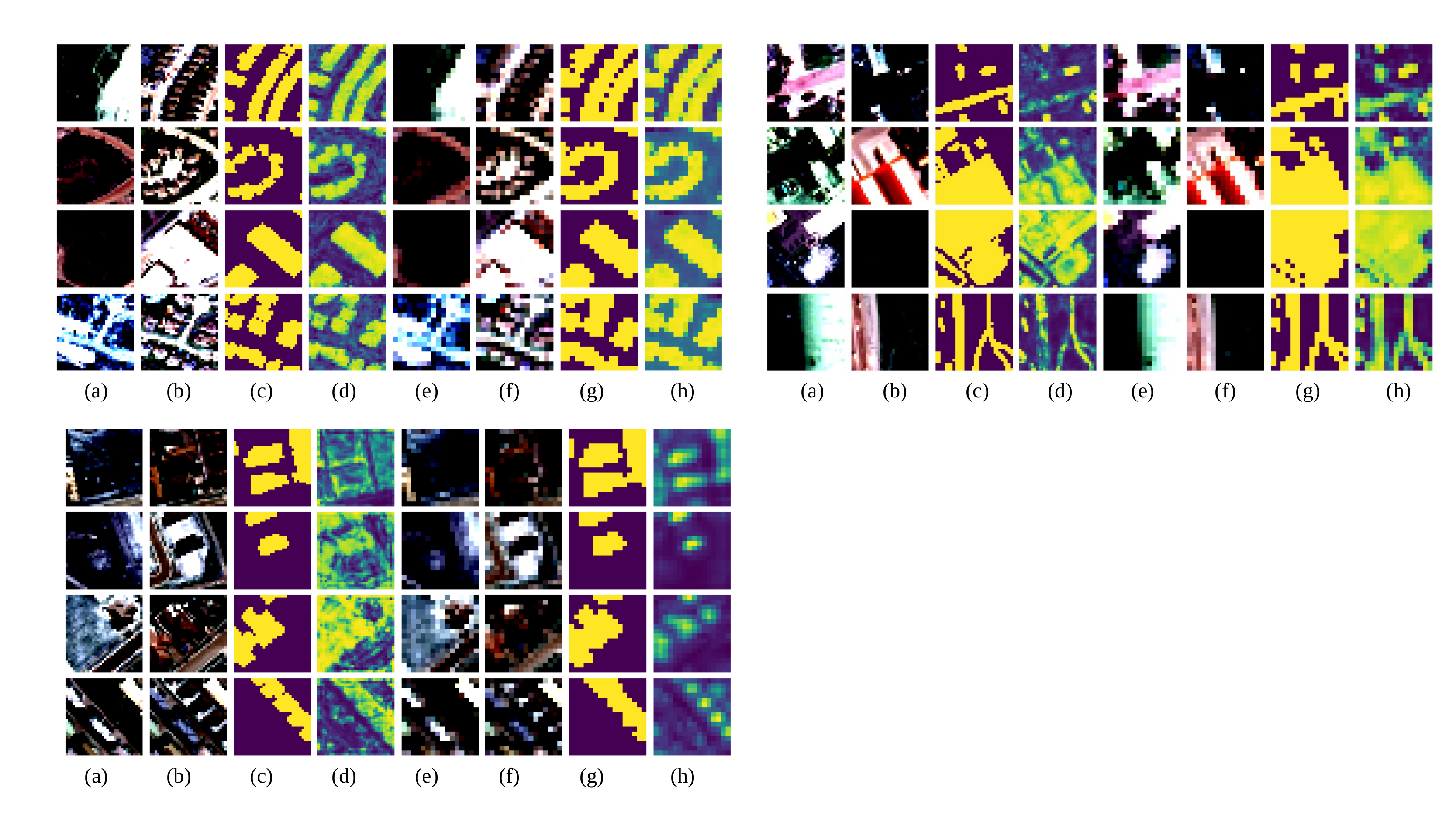}
    \caption{Visualization examples of deep features retrieved by DSFR modules in the WHU-CD dataset. (a) Images from $t_1$  downsampled by 8. (b) Images from $t_2$ downsampled by 8. (c) Ground-truths  downsampled by 8. (d) Retrieved features from the fourth stage. (e) Images from $t_1$  downsampled by 16. (f) Images from $t_2$ downsampled by 16. (g) Ground-truths downsampled by 16. (h) Retrieved features from the fifth stage.}
    \label{fig10}
\end{figure}

\begin{figure}
    \centering
    \includegraphics[width=0.99\linewidth]{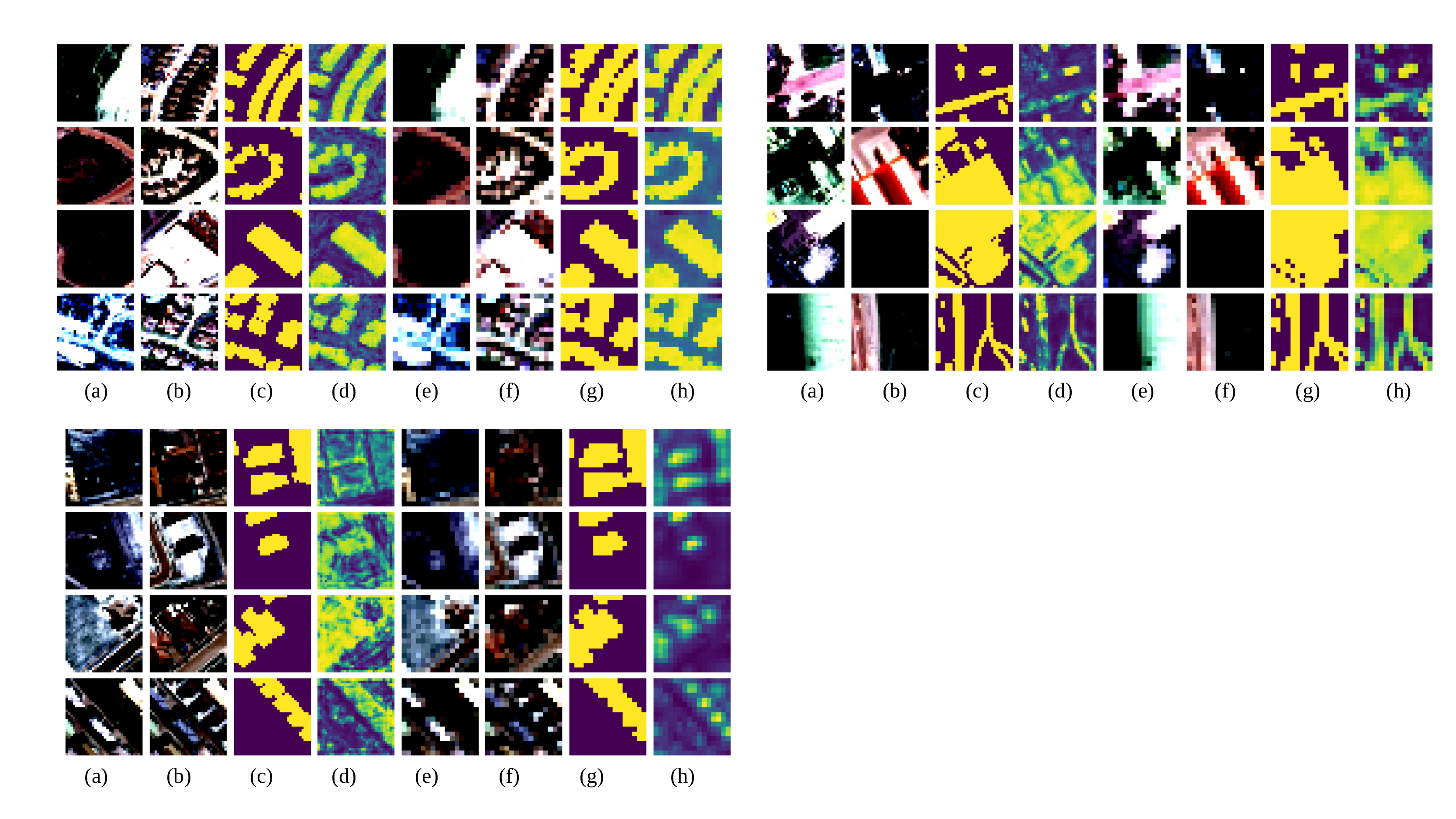}
    \caption{Visualization examples of deep features retrieved by DSFR modules in the CDD dataset. (a) Images from $t_1$  downsampled by 8. (b) Images from $t_2$ downsampled by 8. (c) Ground-truths  downsampled by 8. (d) Retrieved features from the fourth stage. (e) Images from $t_1$  downsampled by 16. (f) Images from $t_2$ downsampled by 16. (g) Ground-truths downsampled by 16. (h) Retrieved features from the fifth stage.}
    \label{fig11}
\end{figure}

As depicted in Figs. \ref{fig9}–\ref{fig11}, significant differences are evident between the downsampling of image pairs across all three datasets, including variations in color, illumination, and even some unchanged objects. However, the deeply supervised feature retrieval module enhances the network's ability to capture textural structures in deeper layers by emphasizing consistency between intermediate change maps and downsampled labels, thereby increasing the network's robustness to real changes. In the LEVIR-CD and CDD datasets, the DSFR modules effectively highlight changes of interest in deeper layers, with intermediate change maps consistent with downsampled ground truths. In the CDD dataset, the dual Hopfield layers also effectively retrieve small objects such as narrow roads and cars. However, in the WHU-CD dataset, the difference feature retrieved in the fourth stage is less accurate. This discrepancy is primarily due to the WHU dataset capturing the target area before and after an earthquake, during which rapid post-disaster reconstruction and urban expansion occurred, resulting in significant changes to new buildings, roads, squares, and vegetation. Consequently, the retrieved difference features differ significantly from the building change labels. Conversely, it is observed that as the network delves deeper, the DSFR module can more accurately locate change information in the fifth stage.

\section{Conclusion}
In this paper, we propose a deep supervision and feature retrieval network for change detection in remote sensing images. Inspired by the modern Hopfield network, we retrieve the different features and aggregate the sequential information in a deeply supervised manner. To reasonably concatenate multi-level features, we adopt a comprehensive fusion strategy into the decoder. Extensive experiments have been conducted on three widely used change detection datasets to validate the effectiveness of the proposed method. In the two building change detection datasets, the proposed model can better preserve the contour of the buildings by avoiding false alarms for dense buildings and controlling miss detection for irregular and independent buildings. For the seasonal variation dataset, the proposed model can accurately capture the real changes among pseudo-seasonal variations and detect small and narrow objects as well. Through an ablation study and sensitivity analysis, we verified the promotion effect of the DSFR module on the backbone network and identified reasonable parametric settings. Finally, visualization examples show the performance of the Hopfield layer on difference feature retrieval and aggregation. Through experiments, we found that the feature retrieval module might fail to locate the change of interests accurately in deeper layers when many unexpected changes occur in the scene. Therefore, we will continue to work on designing appropriate structures to overcome the side effects of pseudo-changes. Additionally, we plan to analyze how different training ratios affect detection accuracy in future work. This analysis may provide deeper insights into the scalability and practicality of our method.

\section*{Acknowledgments}
The authors would like to thank the authors of all comparative methods for sharing their codes, the contributors of the LEVIR-CD, WHU-CD, and CDD datasets, Dr. Yonghao Xu for his valuable comments and discussions, and the Institute of Advanced Research in Artificial Intelligence (IARAI) for its support.

\bibliographystyle{IEEEtran}
\bibliography{ref}


 




\vfill

\end{document}